\documentclass[10pt,conference,a4paper]{IEEEtran}
\usepackage{multirow}
\usepackage{graphicx}
\usepackage{amsmath}
\usepackage{stackrel}
\usepackage{tabularx}
\usepackage[export]{adjustbox}
 
\begin{document}
%
\title{Automated Image Captioning for Rapid Prototyping and Resource Constrained Environments}

\author{ \IEEEauthorblockN{Karan Sharma \;\; Arun CS Kumar \;\; Suchendra M. Bhandarkar }
\IEEEauthorblockA{Department of Computer Science, The University of Georgia, Athens, Georgia 30602-7404, USA}
\IEEEauthorblockA{Emails: \{karan@uga.edu \;\; aruncs@uga.edu \;\; suchi@cs.uga.edu\} }}

\maketitle

\begin{abstract}
Significant performance gains in deep learning coupled with the exponential growth of image and video data on the Internet have resulted in the recent emergence of automated image captioning systems. Ensuring scalability of automated image captioning systems with respect to the ever increasing volume of image and video data is a significant challenge. This paper provides a valuable insight in that the detection of a few significant (top) objects in an image allows one to extract other relevant information such as actions (verbs) in the image. We expect this insight to be useful in the design of scalable image captioning systems. We address two parameters by which the scalability of image captioning systems could be quantified, i.e., the traditional algorithmic time complexity which is important given the resource limitations of the user device and the system development time since the programmers' time is a critical resource constraint in many real-world scenarios. Additionally, we address the issue of how word embeddings could be used to infer the verb (action) from the nouns (objects)  in a given image in a zero-shot manner. Our results show that it is possible to attain reasonably good performance on predicting actions and captioning images using our approaches with the added advantage of simplicity of implementation.  
\end{abstract}


%
\IEEEpeerreviewmaketitle

\section{Introduction}

\noindent
Automated image captioning, i.e., the problem of describing in words the situation captured in an image, is a challenging problem for several reasons. 
The recent significant performance gains in deep learning coupled with the exponential growth of image and video data on the Internet have resulted in the emergence of automated image captioning systems. Although automated image captioning has been addressed in a general sense using a variety of approaches, in this paper we address two specific issues in the context of image captioning:
\begin{enumerate}

\item How does one deploy an automated image captioning system with the goal of simultaneously reducing system development time and CPU execution time? 

\item How does one improve the algorithmic efficiency of the automated image captioning system for implementation on resource-constrained user devices (such as smartphones)? 

\end{enumerate}
\noindent
To address both of the above issues, we draw inspiration from cognitive science. We hypothesize that automated image captioning can be greatly improved by first detecting the most significant (i.e., top objects) objects in an image, and then inferring all the other relevant information from these top objects. In the cognitive science literature, it has been shown that there are inherent properties of the visual and linguistic world that make perception of objects easier than perception of relational categories such as verbs. Objects are seen to be cohesive structures~\cite{Gentner06,Gentner81} and concrete objects, even when malleable, are seen to be held together by percepts or parts which gives the object a property of wholeness. This property of wholeness is not characteristic of relational categories such as verbs.

\begin{figure}
  \centering
  \centerline{\includegraphics[width=8cm]{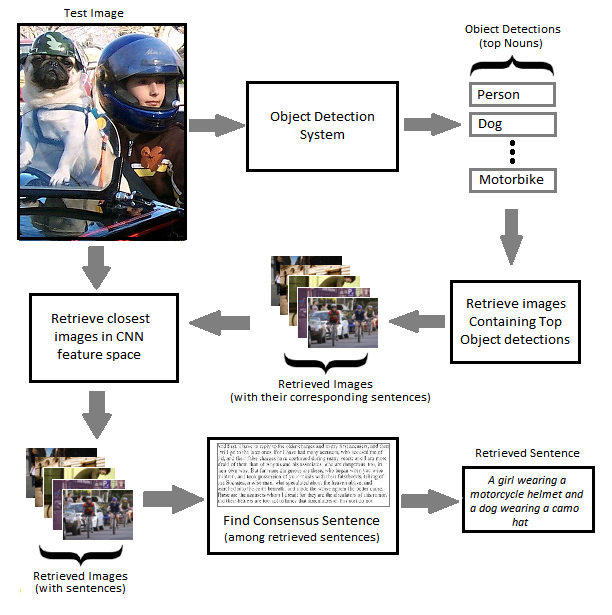}}
   \vspace{-0.2in} 
  \caption{Top-object detections used to drive $k$-NN search.}
\label{fig:sequence1}
\end{figure}

Additionally, objects in an image directly map to concrete entities (i.e.,nouns) in the real world~\cite{Gentner06,Gentner81}. This direct translation between nouns and objects enables the stable assignment of a word to an object in an image. In contrast, relational categories such as verbs tend to describe relations between disjoint entities. Moreover, verbs are also more {\it polysemous} than nouns, in that verbs have more senses of meaning than do nouns~\cite{Gentner81}. For example, the verb {\it running} can be used in a variety of settings - ``John is {\it running} for public office'', and ``John is {\it running} on the field'', use the verb {\it running} in different senses. Likewise, verbs are also {\it mutable} in that their meaning can be adjusted based on the context of sentence~\cite{GentnerFrance88}. For example, in the sentence, ``John is {\it jumping} to a conclusion'',  the verb {\it jumping} adjusts its meaning to the nouns in the sentence. The properties of polysemy and mutability make automated generation of training datasets for verbs a very difficult task.

We believe that the insight provided by the detection of top objects in an image over extraction of other kinds of information is useful in addressing both issues mentioned above. To address the first issue, we use top-object detections to speed up the caption retrieval procedure during automated image captioning. Specifically, we use the detection of the most significant objects in an image (i.e., the top objects) to speed up the $k$-nearest-neighbor ($k$-NN) search for retrieval-based automated image captioning.  We show top-object detection to be a preferable alternative to traditional image captioning methods that employ Locality Sensitive Hashing (LSH) to speed up the $k$-NN search. 
To address the second issue, we propose to identify a significant activity (i.e., verb)  in an image in a zero-shot manner by taking advantage of plausible object-pair detections in an image and word embeddings in a {\it word2vec} representation~\cite{Mikolov131}. The proposed zero-shot approach is particularly useful for implementation on resource-constrained user devices since the need for running computationally intensive activity (i.e., verb) detectors on an image is obviated and the object-pair and verb assignments to an image can be made quickly. 

\section{Related Work}

\noindent
\textbf {Automated image captioning:} Automated image captioning systems have grown in prominence owing, in large part, to tremendous performance gains in deep learning in recent times. Existing automated image captioning systems can be categorized as either generative or retrieval-based. Generative systems involve correctly identifying objects, verbs, adjectives, prepositions or visual phrases in an image and generating a caption from these or directly from the representation of the image~\cite{Donahue14},~\cite{Fang14},~\cite{Karpathy14},~\cite{Krishnamoorthy13},~\cite{Mao14},~\cite{Vinyals14}. 
Retrieval-based approaches~\cite{Devlin15}~\cite{Gentner81}, on the other hand, involve retrieving the most suitable caption from a database of captions and assigning it to an image. All the above approaches address image captioning in a general sense, but our contribution is to address image captioning for two specific situations, one where rapid system development time is needed to deploy image captioning (i.e., the $k$-NN model), and image captioning for resource-constrained devices where algorithmic efficiency is critical (i.e., the second issue).

\noindent \textbf {Action (Verb) recognition}: Action recognition in videos is a relatively easier problem and has been addresed by using various kinds of spatio-temporal descriptors followed by a classifier~\cite{Everts14},~\cite{Peng14},~\cite{Wang13}.  
However, in static images, unlike videos, there is no temporal information. Also, the problem is further exacerbated by the lack of reliable annotation data. Hence, several existing works in action recognition on static images use supervised learning of visual information integrated with linguistic information mined from a text corpus~\cite{Farhadi10},~\cite{Krishnamoorthy13},~\cite{Yang11}. In this paper, we argue that if we know just the objects in an image, we can succesfully use word embeddings to describe the underlying action, even in static images, with a reasonable degree of accuracy. More recently, Jain et. al.~\cite{Jain14} have shown the efficacy of word embeddings for zero-shot action recognition in videos. In contrast, the strength of our approach is that we use word embeddings in a relatively simple manner and yet obtain good results on a challenging set of static images. 

\section{$k$-NN Search Driven by Top-Object Detections} 

Devlin et al.~\cite{Devlin15} have obtained good results for automated image captioning based on $k$-NN image retrieval. Their approach determines the $k$-NN images by computing a measure of image similarity between the test/query image and each of the database images. The test/query image is then assigned the caption obtained by computing the consensus of the $k$-NN image captions. Performing an exhaustive search of the image database to retrieve the $k$-NN images using an image feature-based similarity metric is clearly not a scalable approach.  We show that, in the context of automated image captioning, by detecting all objects in a test image, selecting the top-$n$ objects (where $n$ is a small number) and retrieving all images that contain at least one of these $n$ objects, one can achieve results comparable to those of $k$-NN retrieval via exhaustive search while also obtaining a significant speed up. Fig.~\ref{fig:sequence1} summarizes the proposed approach. 

Techniques such as LSH are traditionally used to speed up $k$-NN image retrieval~\cite{Dong08}. However, the hyperparameter tuning procedure needed to optimize the performance of LSH is non-trivial in terms of computational complexity~\cite{Dong08}, especially in the case of complex applications such as automated image captioning.  Thus, complete automation of LSH for automated image captioning is a challenging task. Implementation of LSH also presents a significant expenditure of system development time, which is an important consideration in real-world situations where rapid prototyping is called for. In contrast, in the proposed approach, tuning the parameters of a support vector machine (SVM)-based classifier for object detection/recognition is much simpler and yields readily to automation. 

Although running various object (i.e., noun) detectors on the test image imposes a computational overhead, it is offset by the following considerations: (a) the space of objects (i.e., nouns) is bounded,
(b) sliding windows are not used during the object detection procedure 
and, (c) real-time object detection and classification is limited only to a single image (i.e., the test image). The computational overhead of object detection in the test image is also offset by: (a) the resulting speedup over $k$-NN image retrieval via exhaustive search and, (b) savings in development time compared to the scenario wherein $k$-NN image retrieval is optimized using LSH.  Additionally, the proposed approach also results in significant savings in CPU execution time as shown in Table~\ref{tab:table1}. 

\noindent
\textbf{Training:}  We train an SVM-based object detector/classifier for each of the 80 annotated object categories in the Microsoft (MS) COCO dataset~\cite{Lin14}. The inputs to the SVM-based classifiers are the VGG-16 {\it fc-7} image features~\cite{Simonyan14} 
extracted using the Matconvnet package~\cite{Vedaldi14}. In addition, we store each training image in the MS COCO dataset and its accompanying sentences  (5 sentences per image) in our database. We treat these sentences as ground truth captions for the corresponding training image. For testing purposes, we consider the MS COCO validation set consisting of close to 40,000 images.

\noindent
\textbf{Testing:} For each test image in the MS COCO validation set, we run all the 80 object detectors on the test image. We select the top-$n$ objects from all the detected objects in the image. In our current implementation $n=5$. The detected top objects are the ones that are deemed to possess the highest probability of occurrence in the image. The probability of occurrence of an object in the image is computed by mapping the classification confidence value generated by the SVM classifier for that object to a corresponding probability value using Platt scaling~\cite{Platt99}. 

From the training dataset, we retrieve all images that contain at least one of the top-$n$ objects detected in the previous step, using the corresponding ground truth captions, i.e., a training image is retrieved if at least one of its associated ground truth captions contains a noun describing the object under consideration. Using the cosine distance between the {\it fc-7} features of each retrieved image and the test image, we select the $k$-NN images for further processing. 

In the current implementation we have chosen $k=90$ as recommended by~\cite{Devlin15}. Since each of the $k$-NN images has 5 associated sentences (captions), we have a total of $5k$ potential captions for the test image. We determine the centroid of the $5k$ potential captions and deem it to represent the consensus caption for the test image.  The consensus caption is then assigned to the test image in a manner similar to~\cite{Devlin15}. The BLEU measure is used to evaluate the similarity (or distance) between individual captions and to determine the centroid of the $5k$ potential captions. We also implement $k$-NN image retrieval using exhaustive search~\cite{Devlin15} and compare the CPU execution time of the proposed approach with that of $k$-NN image retrieval using exhaustive search for 2000 random images .

\noindent
\textbf{Results:} As shown in Table~\ref{tab:table1}, the proposed $k$-NN retrieval driven by top-object detections and the standard $k$-NN retrieval that employs exhaustive search yields very similar results when the BLEU and CIDEr~\cite{Vedantam14} similarity metrics are used to compare captions. 
Additionally, the proposed approach is seen to yield significant gains in CPU execution time when compared to $k$-NN image retrieval using exhaustive search.
Essentially, the proposed $k$-NN retrieval technique driven by top-object detections is observed to provide an attractive alternative to LSH for the purpose of speeding up $k$-NN image retrieval in the context of automated image captioning. In spite of its success, optimal employment of LSH entails fine tuning of its hyperparameters, a potentially cumbersome and time-consuming procedure, especially when dealing with large datasets ~\cite{Dong08},~\cite{Slaney12}. 

Additionally, the experimental results show that other relevant forms of information such as verbs (i.e., actions) and adjectives (i.e., attributes) co-occur reliably with nouns (i.e., objects) in an image. Table~\ref{tab:table2} shows the precision for several verbs in the retrieved captions. The precision results in Table~\ref{tab:table2} suggest, that the associated verbs can be reliably retrieved from just the top-nouns. The resulting precision in the range 30\%--50\% is encouraging because there could be multiple verbs that could describe a situation in an image. For instance, {\it riding} and {\it driving} or, {\it eating} and {\it grazing} could be deemed as reliable verbs that describe the same situation.  Hence, the top objects in an image could be used as a prior for predicting or retrieving other relevant information in an image. 

\begin{table}
 \label{tab:table1} 
 \footnotesize
 \caption{Comparison of image captioning results obtained using the proposed approach based on top-objects detection-driven $k$-NN retrieval (Obj-$k$-NN) versus $k$-NN retrieval based on exhaustive search (Exh-$k$-NN).}
 \centering
\scriptsize
    \begin{tabular}{| c | c | c | c | c | c | c | c}
    
    \hline
   \ & BLEU$_1$ & BLEU$_2$ & BLEU$_3$ & BLEU$_4$ & CIDEr & CPU time\\ \hline    
    {Exh-$k$-NN} & 65.6\% & 47.4\% & 34\% & 24.7\%   & 0.70\% & 2.5e+04s\\ \hline
    {Obj-$k$-NN} & 64.6\% & 46.2\% & 32.8\% & 23.6\% &  0.68\% & 1.17e+04s\\ \hline    
    \end{tabular}    
\end{table}

\begin{table}
 \label{tab:table2}  
\footnotesize
 \caption{Precision of most frequently occurring verbs in the MS COCO dataset that were extracted from captions obtained via $k$-NN retrieval driven by top-object detections.}
 \centering
\footnotesize
    \begin{tabular}{| c | c | c }
    \hline
    \ Verb & Precision  \\ \hline    
    \ $Sit$ &  0.47 \\ \hline
    \  $Stand$ &  0.48   \\ \hline    
        \ $Hold$ &  0.39 \\ \hline
    \  $Ride$ & 0.45   \\ \hline 
        \ $Walk$ &  0.3 \\ \hline
    \  $Play$ & 0.49   \\ \hline  
    \end{tabular}   
\end{table}

\section{Top-object Detection-driven Verb Prediction Model}

\begin{figure}
  \centering
  \centerline{\includegraphics[width=9cm]{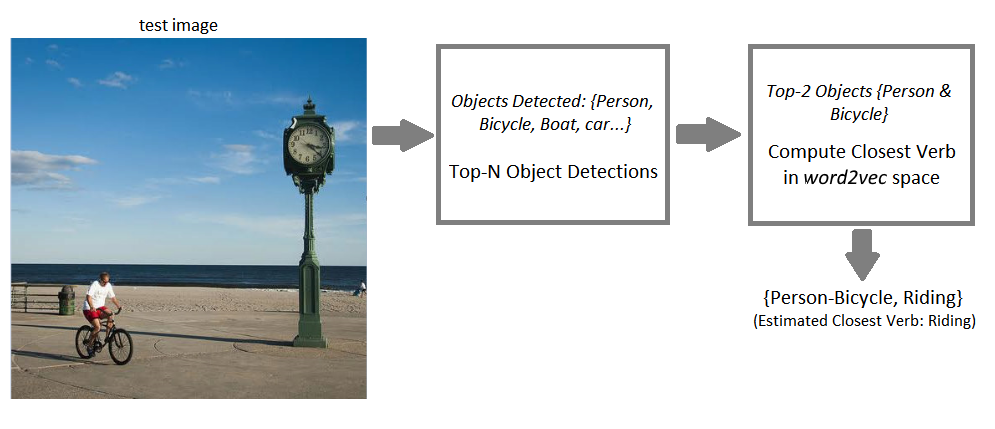}}
   \vspace{-0.1in} 
  \caption{Outline of the proposed top-object detection-driven verb prediction model}
\label{fig:topobject}
\end{figure}

In this approach, as depicted in Fig.~\ref{fig:topobject}, we detect the top objects in an image, identify the most plausible two objects (i.e., object pair) in the image, and then assign the most meaningful action (verb) to this object pair. Unlike the $k$-NN approaches described in the previous section, this approach could be useful in situations where we do not have a very large and diverse training dataset. We show that by learning an appropriate {\it word2vec} representation~\cite{Mikolov13}, the assignment of a verb (activity) to an object pair can be achieved both accurately and quickly. 

With the recent proliferation of resource-constrained devices that constitute the Internet-of-things (IoT), it is important to have image analysis and retrieval techniques that could provide significant algorithmic time gains. Hence, by recognizing the {\it object-pair} and associated {\it verb} in a time-efficient manner, one could describe the crux of the story underlying an image even in resource-constrained environments. Whereas feature extraction and object detection/classification are unavoidable in an automated image captioning system, we believe that when inferring a relational category, significant algorithmic time gains can be achieved if we can reliably infer a verb from its associated objects in constant (i.e., $O(1)$) time.

The proposed verb prediction model uses the {\it word2vec} representation scheme~\cite{Mikolov131} which is based on the embedding of a word in a hypothetical low-dimensional vector space. In the {\it word2vec} representation scheme, each word is mapped to a point in the hypothetical vector space such that words that have similar meanings tend to be proximal in this vector space. In our case, we intend to capture the relationships between nouns and verbs, for example {\it pizza} and {\it eat}, and noun-pairs and verbs, for example, {\it Person-dog} and {\it walk}.

It is interesting to examine why the {\it word2vec} representation is able to learn the word embeddings that capture the relations between nouns and verbs, or between noun-pairs and verbs. Note that the {\it word2vec} representation groups semantically similar words into proximal regions in the hypothetical vector space, i.e., words that are similar in meaning such as {\it beautiful} and {\it pretty} are mapped to proximal points in the hypothetical vector space. In this sense, the {\it word2vec} representation treats synonymy, not as a binary concept, but rather one that spans a continuum. However, we hypothesize that even when the words are not obviously synonymous or similar in meaning, the distance between their corresponding points in the vector space can still tell us something significant about their relationship.

For the sake of clarification, consider the following example. Assume that we are given a collection of the following four sentences:  

\noindent {\it ``A person is driving a car on the road". ``A car is passing a truck on the road". ``A car is parked on the road". ``A person is driving a truck".}

In the above sentences, the context of the noun {\it car} is \{{\it person, road, truck, driving}\} whereas the context of the verb {\it driving} is \{{\it person, road, truck, car}\}. As the contexts of {\it car} and {\it driving}  are very similar, {\it word2vec} will place the embeddings of {\it car} and {\it driving} in close proximity in the vector space although  {\it car} and {\it driving} are strictly not synonymous words. Based on context, among all verbs, the verb {\it driving} will tend to be closer to the noun {\it car} based on their respective embeddings in the vector space. For a more rigorous treatment of why the {\it word2vec} embedding tends to capture such linguistic regularities the interested reader is referred to~\cite{Pennington14}.

The problem of determining the closest verb to two given top nouns can be stated as follows. Given a set of verbs $V$ and top nouns $n_1$ and $n_2$, the closest verb from the set $V$ to the top nouns $n_1$ and $n_2$ is given by:
\begin{equation}
\underset{i}{\arg\max}\:\:\{SIM (v_{i}, n_{1})+SIM (v_{i}, n_{2})\}
\label{eq:sim1} 
\end{equation}
where $SIM(v_{i}, n_{1})$ and $SIM(v_{i}, n_{2})$ are the cosine similarities of verb $v_i$ to noun $n_1$ and $n_2$ respectively.

One of the problems with above formulation is that certain nouns such as {\it person} and {\it apple}, when considered independently, may have multiple verbs that are proximal in vector space. For example,  {\it person} and {\it apple}, when considered independently, may be proximal to multiple verbs such as  {\it sit}, {\it hold}, {\it sleep} and so on. Simple addition of the cosine similarities as shown above does not bias the verb prediction towards {\it eat} when {\bf both} the nouns {\it person} and {\it apple} are present in the same sentence. To circumvent the above problem, in the sentence database accompanying the MS COCO training dataset, we append each sentence with all object-pairs occurring in that sentence. In other words, we identify all the nouns in a sentence and form all pairs of these nouns before appending them to the sentence. For example, given a sentence
\noindent
{\it ``Person is eating an apple sitting on the table."}, 

\noindent
we convert the sentence into following two sentences: 

\noindent
{\it ``Person is eating an apple sitting on the table apple-person."}

\noindent
{\it ``Person is eating an apple sitting on the table person-table."} 

\begin{figure}
  
  \includegraphics[width=9cm]{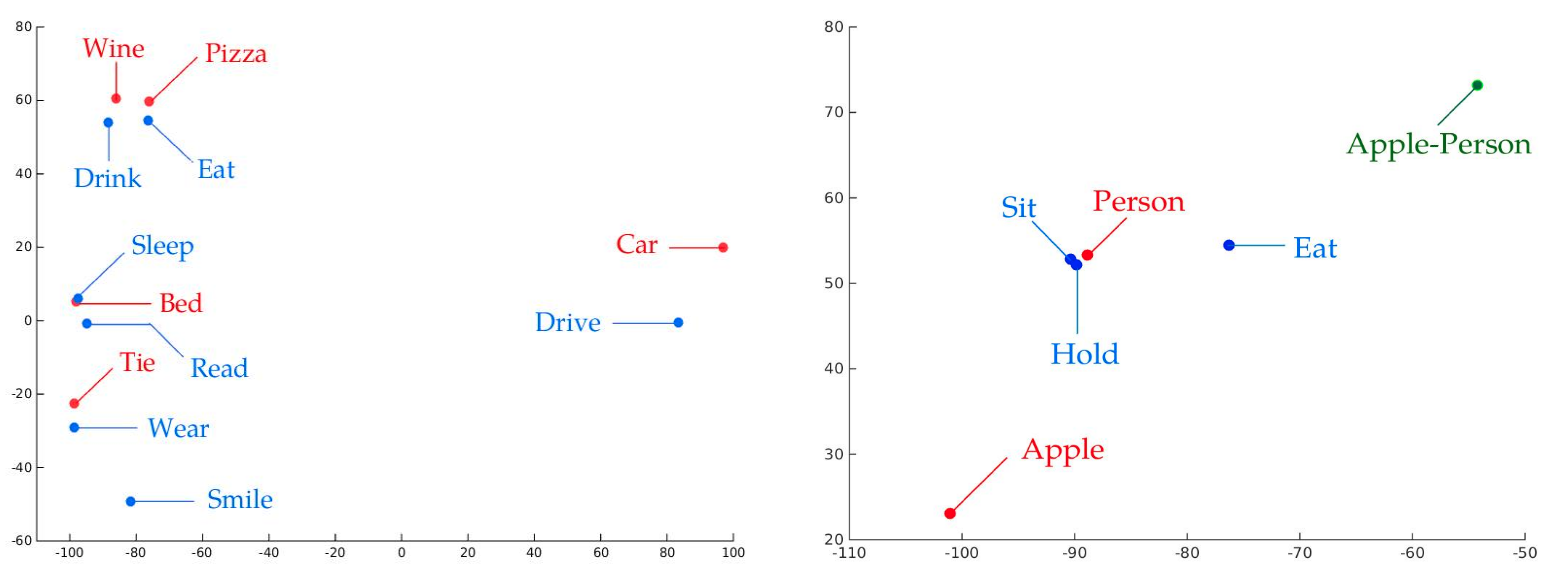}
   \vspace{-0.2in} 
  \caption {Visualization of word embeddings is 2D space using {\it t-SNE} dimensionality reduction~\cite{Vander08}.  Left:  Most verbs tend to occur closer to their attached nouns. Right: Appended nouns ({\it apple-person}) occur nearer to verb {\it eat} than individual nouns {\it apple} and {\it person}.}
\label{fig:sequence3}
\end{figure}

This simple preprocessing step potentially captures the dependence between the noun-pairs and the verb. Next, we train the {\it word2vec} model on the modified sentence database. After the model is trained, it will have learned the verb that defines a relationship between a pair of objects. Fig.~\ref{fig:sequence3} clarifies the above argument using the projection of these word embeddings in a 2Dspace using {\it t-SNE} dimensionality reduction~\cite{Vander08}.

More formally, given the set of verbs $V$, and noun-pair $n_{12}$, the closest verb in $V$ to the noun-pair is $n_{12}$ is given by: 
\begin{equation}
\underset{i}{\arg\max}\:\:SIM({v_{i}, n_{12}})
\label{eq:sim2} 
\end{equation}
where $SIM(v_{i}, n_{12})$ is the cosine similarity between the verb $v_i$ and noun-pair $n_{12}$ . 

In the above model, once the necessary steps of feature extraction and object detection are performed, verb prediction for a given noun-pair can be achieved in $O(1)$ time. During testing, the top verbs can be easily retrieved in $O(1)$ time using an appropriate hash data structure once the top-2 nouns (objects) in the test image are detected. Since the verb is detected in a zero-shot manner, the computational expense of running verb detectors on the image is obviated. After the {\it word2vec} model is trained on the modified sentence dataset, we choose plausible verbs that are closest in distance to each object-pair and store the verbs in the database along with the object-pair.

\noindent
\textbf{Training:} The {\it word2vec} model is trained with a window size of 10 using the implementation of Rehurek and Sojka~\cite{Rehurek10}. This results in a 300-dimensional vector for each word using the skip-gram model for {\it word2vec} ~\cite{Mikolov13}. In the skip-gram approach, the input to the deep-learning neural network (DLNN) is the word, and the context is predicted from the word. 
For example, given the contextual input {\it eat}, the model will predict \{{\it person}, {\it pizza}\}.  To train skip-gram, given a sequence of words $w_1, w_2, w_3,....w_T$, we maximize the following objective function:
\begin{equation}
\frac{1}{T}\stackrel[t=1]{T}{\sum}\underset{-c\leq j\leq c,j\neq0}{\sum}\log p(w_{t+j}|w_{t})
\label{eq:skip} 
\end{equation}
where $c$ is the context parameter that specifies the number of words to be predicted from a given word~\cite{Mikolov132}. The term $p(w_{t+j}|w_{t})$ signifies the prediction probability of the context given the word. The stochastic gradient descent algorithm is used for training the skip-gram model. More details on the skip-gram architecture can be found in ~\cite{Mikolov132}. After the model is trained, we obtain the word embeddings corresponding to each word in the dataset. 


The nouns within each sentence are converted to noun-pairs and appended to the end of the sentence, as explained previously. We also perform a couple of additional preprocessing steps on the entire data.  First, we stem each word using Porter's stemmer; for instance, {\it driving} and {\it drive} are both converted to {\it driv}. 

Additionally, words synonymous to {\it person} such as {\it human}, {\it woman}, {\it boy},  {\it girl}, {\it people} etc. are converted to {\it person}. Currently, we try to infer only the most frequently occurring verbs in the MS COCO dataset, i.e., we select the top-$n$ (where $n=51$) most frequently occurring verbs in the MS COCO dataset. The top verbs in the COCO dataset are obtained by parsing the training captions using the Stanford parser. 
The 40,000 images in the COCO validation set are split into two subsets, each containing 20,000 images; one subset is used for validation of the hyperparameter tuning procedure and the other subset for testing. The validation subset is used to learn the hyperparameters and also the other required parameters for the skip-gram model. 

\noindent
\textbf{Testing:} Given a test image, we run all the object detectors on it and select the top-2 highly probable object detections as candidate objects. For this object-pair, we use the {\it word2vec} model to get the closest verbs. 

For each test image, we recognize an object-pair, and predict the plausible verbs in the image. Among the comparison measures introduced below, all except one, predict two plausible verbs in an image. If any one of the predicted verbs occurs in any of the ground truth captions of an image, we regard the prediction as accurate. In addition, even if there are multiple verbs in the ground truth captions for a particular image, intuitively, we just need to find one accurate verb to describe the crux of a story in an image. For example, if the two ground truth captions for a particular image are {\it Person is riding a motorcycle} and {\it Person is driving a motorcycle}, it would suffice to just get one of the two verbs {\it riding} or {\it driving} correct.  Hence, when computing the prediction accuracy, we aim to get just one verb correct in the ground truth captions.

We report results on the subsets $S_1$  and $S_2$ of the validation set of the MS COCO dataset, which we use for testing purposes. $S_1$ is a subset of the validation dataset wherein the ground truth captions have at least two objects from the annotated noun set, and at least one verb from top-51 most frequently occurring verbs in the COCO dataset. $S_2$ is a subset of $S_1$ wherein the top-2 objects have been correctly detected in an image. These results are used to show how effectively a verb is inferred after the object-pair is correctly detected in an image. We compare the results of the proposed scheme using six possible evaluation scenarios:

\noindent {\it Random Baseline (Rand):} where the two verbs are generated randomly for the top-noun detections in an image. 

\noindent {\it 1-Obj Baseline (1-Obj):} where the top-most object (object with highest probability) is used to predict top-2 verbs in an image using word embeddings. The top-2 verbs that are closest in distance to this top-object are selected.

\noindent {$VD_1$:}  where the top-2 closest verbs are the one with the lowest mean distance from the top two object detections measured using equation (\ref{eq:sim1}). 

\noindent {$VD_2$:}  where the top-2 closest verbs are the one with the lowest distance from the appended noun-pair  measured using equation (\ref{eq:sim2}). 

\noindent {$VD_3$:}  where the top two verbs are assigned as follows: if the closest verb determined using equations (\ref{eq:sim1}) and (\ref{eq:sim2}) is the same, we assign this verb to an image, and the second closest verb is assigned using equation (1). Otherwise, one of the top two verbs is assigned using equation (\ref{eq:sim1}) and the other using equation (\ref{eq:sim2}).  

\noindent {$VD_4$:} where the verbs are assigned using set union between the top three verbs determined using equations (\ref{eq:sim1}) and (\ref{eq:sim2}).  

A comparison between the six possible evaluation scenarios is shown in Table~\ref{tab:table3}. 

\begin{table}
\footnotesize
 \label{tab:table3}    
 \caption{Comparison of the proposed model with a random baseline and the 1-object baseline. DS denotes the data subset and accuracy denotes the percentage of images where both \textit{object-pair and verb} are correctly predicted.}
 \centering
\footnotesize
    \begin{tabular}{| c | c | c | c | c | c | c | c}
    \hline
    \ DS & Rand & 1-Obj  & $VD_1$ & $VD_2$ & $VD_3$ & $VD_4$ \\ \hline    
    \ $S_1$ &  1\% & 35.2\% & 36.9\% & 31.4\% & 32.8\% & 56.63\%  \\ \hline
    \  $S_2$ & 9\% &  45.81\% & 53.43\% & 57.74 \% & 52.35 \% & 73.09\% \\ \hline     
    \end{tabular} 
\end{table}

\noindent
\textbf{Results:} The results in Table~\ref{tab:table3} lend support to our claim that top-object detections could be used to infer other information in an image such as verbs. Once we know the plausible object pairs, we can infer the corresponding verb in $O(1)$ time. It is evident from Table ~\ref{tab:table3} that if we know the pair of objects in an image, we can predict the verb with reasonable accuracy.

\begin{figure}
  \centering
  \centerline{\includegraphics[width=9.5cm]{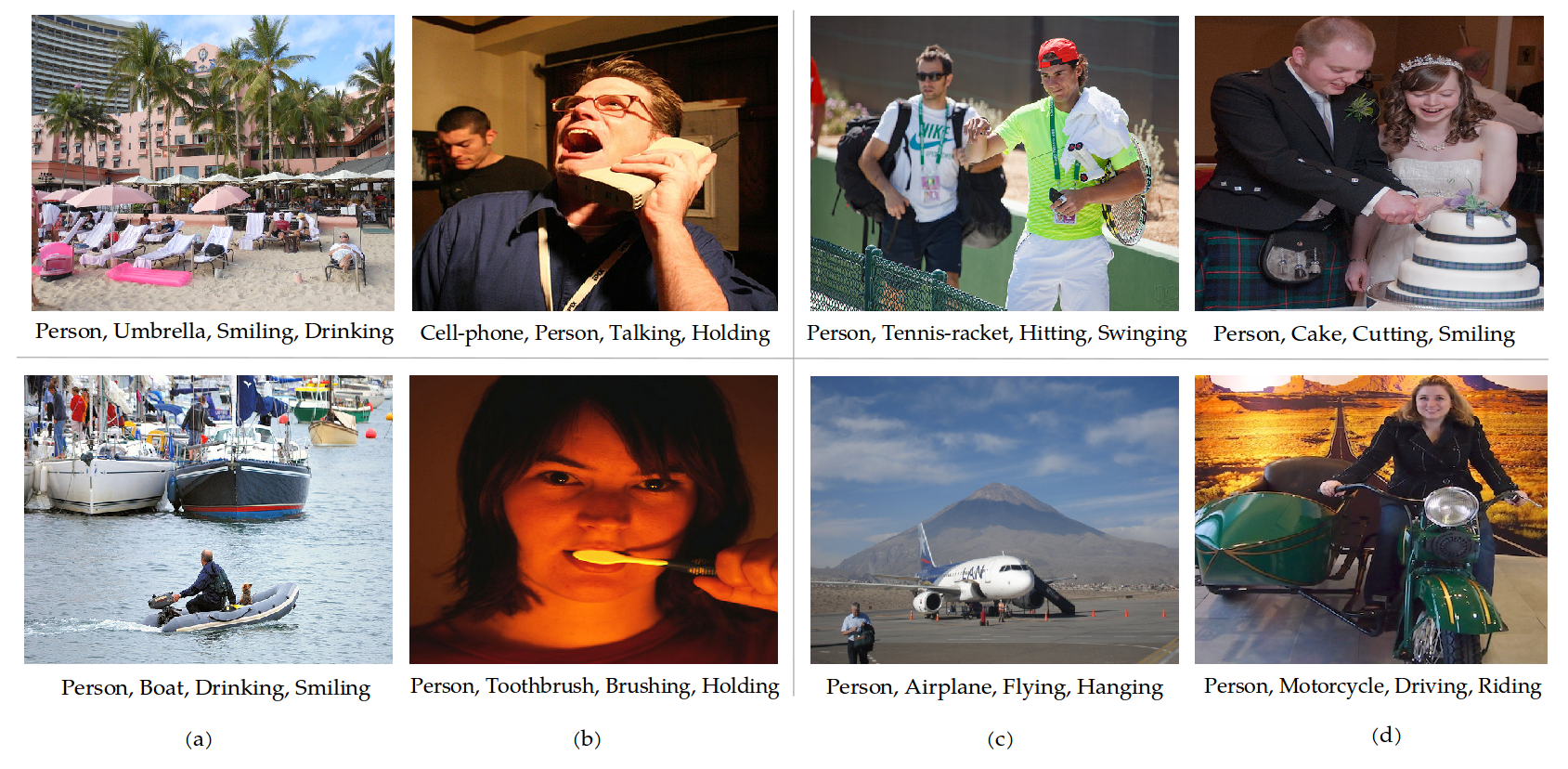}}
   \vspace{-0.1in} 
  \caption{Qualitative Results for Verb Prediction model. (a) $VD_1$: Bad; (b) $VD_1$: Good; (c) $VD_2$: Bad; (d) $VD_2$: Good}
\label{fig:sequence2}
\end{figure}

If the object-pair is correctly recognized in an image, the results in the case of $VD_2$ are slightly better than those in the case of $VD_1$ and $VD_3$. Hence the proposed technique of appending the object-pair to the end of the sentence does provide a non-trivial benefit for verb prediction. However, if the object-pair is not correctly identified in an image, then $VD_1$ outperforms $VD_2$ and $VD_3$. In other words, finding the closest verb by computing the mean distance to the top-2 nouns is better than using the joint object-pair when at least one of the objects is incorrectly detected. the qualitative results for $VD_1$ and $VD_2$ are shown in Fig.~\ref{fig:sequence2}. 

In the case of $VD_3$, where we try to get the combined benefits of $VD_1$ and $VD_2$ the results were inferior to those of both $VD_1$ and $VD_2$. This could be attributed to the fact that results in the case of $VD_1$ and $VD_2$ had only a marginal quantitative difference. This appeared to suggest that to get actual benefits of both $VD_1$ and $VD_2$, we may need to predict more than 2 verbs.  Hence, in the case of $VD_4$, where we predict multiple verbs using both $VD_1$ and $VD_2$ we are far more successful in getting at least one verb correct. We believe that besides predicting multiple verbs, there are a couple of other reasons for the relative success of $VD_4$: there are situations where $VD_1$ will be successful, and there are situations where $VD_2$ will be successful;  $VD_4$ denotes the best of both worlds where $VD_1$ corrects and compensates for weakness of $VD_2$ and vice versa. Also, the high accuracy of $VD_4$ suggests if we try to predict a few more verbs (of the order of 3-6), than there is a very high probability of getting at least one of them correct. 
Overall, from the results it is clear that just detecting the objects in an image is enough to predict a verb (action) in an image with competitive accuracy. Our proposed approaches successfully beat the baseline results lending support to our claim. 

\section{Conclusions}

We have addressed two major issues in the design of automated image captioning system: (a) reduction of  programmers' time and CPU time for rapid deployment, and (b) reduction in algorithmic complexity for implementation on resource-constrained user devices. Our approach has an advantages of ease of use and scalability for automated image captioning and action recognition while delivering competitive results. Future work will entail more sophisticated use of word embeddings to handle even more complex situations. 

\begin{thebibliography}{11}

\bibitem{Devlin15}
Devlin, J.  et al.  (2015). Exploring Nearest Neighbor Approaches for Image Captioning. arXiv preprint arXiv:1505.04467. 

\bibitem{Devlin151}
Devlin, J. et al.  (2015). Language Models for Image Captioning: The Quirks and What Works. {\it Proc. ACL 2015}.

\bibitem{Donahue14}
Donahue, J. et al. (2014). Long-term recurrent convolutional networks for visual recognition and description. {\it Proc. IEEE CVPR}.

\bibitem{Dong08}
Dong, W. et al. (2008). Modeling lsh for performance tuning. {\it Proc. ACM Conf. Info. \& Know. Mgmt.} October, pp. 669-678. 

\bibitem{Everts14}
I. Everts et al. (2014)  Evaluation of color spatio-temporal interest points for human action recognition. {\it IEEE Trans. Img. Process.} 23(4), pp. 1569-1580.

\bibitem{Fang14}
Fang, H. et al. (2014). From captions to visual concepts and back. arXiv preprint arXiv:1411.4952. 

\bibitem{Farhadi10}
Farhadi, A. et al.  (2010). Every picture tells a story: Generating sentences from images. {\it Proc. Eur. Conf. Comp. Vis. (€"ECCV 2010)},  pp. 15-29. 

\bibitem{Gentner06}
Gentner, D. (2006). Why verbs are hard to learn. {\it Action meets word: How children learn verbs}, pp. 544-564. 

\bibitem{GentnerFrance88}
Gentner, D. \& France, I. M. (1988). The verb mutability effect: Studies of the combinatorial semantics of nouns and verbs. {\it Lexical Ambiguity Resolution: Perspectives from Psycholinguistics, Neuropsychology, and Artificial Intelligence}, pp. 343-382. 

\bibitem{Gentner81}
Gentner, D. (1981). Some interesting differences between verbs and nouns. {\it Cognition and Brain Theory}, 4(2), pp. 161-178. 

\bibitem{Harris54}
Harris, Z. (1954). Distributional structure. {\it Word}, 10(23): 146-162.

\bibitem{Jain14}
Jain, M. et al. (2015) Objects2action: Classifying and localizing actions without any video example. {\it Proc. IEEE Intl. Conf. Comp. Vis. (ICCV 2015)}. 

\bibitem{Karpathy14}
Karpathy, A. \& Fei-Fei, L. (2015). Deep visual-semantic alignments for generating image descriptions. {\it Proc. IEEE Conf. Comp. Vis. Patt. Recog. (CVPR 2015)}. 

\bibitem{Krishnamoorthy13} 
Krishnamoorthy, N. et al. (2013). Generating Natural-Language Video Descriptions Using Text-Mined Knowledge. {\it Proc.  AAAI}, Vol. 1, pp. 2. 

\bibitem{Lin14}
Lin, T. Y. et al. (2014). Microsoft COCO: Common objects in context. {\it Proc. Eur. Conf. Comp. Vis. (€"ECCV 2014)}, pp. 740-755. 

\bibitem{Mao14}
Mao, J. et al. (2014). Explain images with multimodal recurrent neural networks. {\it Proc. NIPS 2014}. 

\bibitem{Mikolov13}
Mikolov, T. et al. (2013). Distributed representations of words and phrases and their compositionality. {\it Proc. NIPS 2013}, pp. 3111-3119. 

\bibitem{Mikolov131}
Mikolov, T. et al. (2013). Efficient estimation of word representations in vector space. {\it Proc. Intl. Conf. Learn. Rep. (ICLR 2013)}.

\bibitem{Mikolov132}
Mikolov, T. et al. (2013). Distributed representations of words and phrases and their compositionality. {\it Proc. NIPS 2013}, pp. 3111-3119.

\bibitem{Peng14}
Peng, X. et al. (2014) Action recognition with stacked Fisher vectors. {\it Proc. Eur. Conf. Comp. Vis. (€"ECCV 2014)}.

\bibitem{Pennington14}
Pennington, J. et al. (2014) Glove: Global Vectors for Word Representation. {\it Proc. Conf. EMNLP}. Vol. 14.

\bibitem{Platt99}
Platt, J. (1999) Probabilistic outputs for support vector machines and comparisons to regularized likelihood methods. {\it Advances in Large Margin Classifiers}, Vol.10(3), pp. 61-74.

\bibitem{Rehurek10}
Rehurek, R. \& Sojka, P. (2010). Software framework for topic modeling with large corpora. {\it Proc. LREC 2010 Wkshp. New Challenges for NLP Frameworks}. Valletta, Malta, pp. 46-50. 

\bibitem{Simonyan14}
Simonyan, K. \& Zisserman, A. (2014). Very deep convolutional networks for large-scale image recognition. {\it Proc. Intl. Conf. Learn. Rep. (ICLR 2014)}.

\bibitem{Slaney12}
Slaney, M. et al. (2012). Optimal parameters for locality-sensitive hashing. {\it Proc. IEEE}, 100(9), pp. 2604-2623.

\bibitem{Vander08}
Van der Maaten, L., and Hinton, G. (2008). Visualizing data using {\it t-SNE}. Jour. Mach. Learn. Res., 9(2579-2605), 85.

\bibitem{Vedaldi14}
Vedaldi, A. \& Lenc, K. (2015). MatConvNet-convolutional neural networks for MATLAB. {\it Proc. ACM Conf. Multimedia Systems (MMSys 2015)}.

\bibitem{Vedantam14}
Vedantam, R. et al. (2015). Cider: Consensus-based image description evaluation. {\it Proc. IEEE IEEE Conf. Comp. Vis. Patt. Recog. (CVPR 2014)}. 


\bibitem{Vinyals14}
Vinyals, O. et al. (2014). Show and tell: A neural image caption generator. {\it Proc. IEEE IEEE Conf. Comp. Vis. Patt. Recog. (CVPR 2014)}. 

\bibitem{Wang13}
Wang, H. \& and Schmid, C. (2013) Action recognition with improved trajectories.  {\it Proc. IEEE Intl. Conf. Comp. Vis. (ICCV 2013)}. 


\bibitem{Yang11}
Yang, Y. et al. (2011). Corpus-guided sentence generation of natural images. {\it Proc. Conf. EMNLP}, pp. 444-454. 



\end{thebibliography}
\end{document}